# A technical note for the 91-clauses SAT resolution with Indirect QAOA based approach

G. Fleury and P. Lacomme

*Université Clermont-Auvergne, Clermont-Auvergne-INP, LIMOS (UMR CNRS 6158),*
*1 rue de la Chebarde,*
*63178 Aubière Cedex, France*
gerard.fleury@isima.fr, philippe.lacomme@isima.fr

**Abstract.** This paper addresses the resolution of the 3-SAT problem using a QAOA-like approach. The chosen principle involves modeling the solution ranks of the 3-SAT problem, which, in this particular case, directly represent a solution. This results in a highly compact circuit with few gates, enabling the modeling of large-sized 3-SAT problems. Numerical experimentation demonstrates that the approach can solve instances composed of 91 clauses and 20 variables with an implementation based on Qiskit.

## 1. Introduction

The paper assumes the definitions and notations standard in propositional satisfiability (SAT) and maximum satisfiability (MaxSAT) as established by (Biere et al., 2009). Maximum satisfiability (MaxSAT) represents an optimization variant of SAT, wherein the goal is to find an optimal truth assignment instead of merely a satisfying one. Within MaxSAT, the optimization objective is determined by a collection of weighted soft clauses. The objective value of a truth assignment is the total sum of the weights associated with the soft clauses it satisfies. Moreover, the MaxSAT problem may encompass hard clauses that necessitate satisfaction from the truth assignment. Many optimization problems can naturally be formulated as instances of MaxSAT.

The 3-SAT problem involves determining whether a given 3-CNF (Conjunctive Normal Form) has a satisfying assignment. An outstanding and celebrated open question pertains to whether solving $k$-SAT for $k \geq 3$ necessitates exponential time.

While all NP-complete problems share equivalence in terms of the existence of polynomial-time algorithms, there exists considerable diversity in the worst-case complexity of known algorithms for resolving 3-SAT. Examining algorithms designed for 3-SAT resolution necessitates an exploration of the range in worst-case complexity. The SAT problem has been established as NP-complete (Cook, 1971), implying that algorithms for solving SAT instances may require exponential time in the worst case. Consequently, various techniques have been proposed to enhance the efficiency of SAT solvers, categorizing them into two main groups: Conflict Driven Clause Learning solvers (CDCL) (Marques-Silva and Sakallah, 1999) and Stochastic Local Search solvers (SLS). CDCL approaches are backtracking approaches taking advantages to heuristic/optimization pruning pruning methods. SAT problems often exhibit symmetries that can be pre-computed to improve the original SAT problem with symmetry breaking clauses.

A large number of classical algorithms have been introduced to efficiently solve instances of the SAT problem and a large number of these algorithms are heuristics which efficiency strogly depend on the density $\alpha = n / k$ of the SAT formulas where $k$ is the number of variables in clause. For density $\alpha \geq 2^m \log 2$, this problem is unsatisfiable with high probability (Franco and Paull, 1983) and conversely, for a density of $\alpha < 2\ k/k$, the SAT problem is satisfiable with a high probability (Ming-Te and Franco, 1990; Chvátal and Reed, 1992).

Within the realm of computational complexity theory, the Maximum Satisfiability Problem (MAX-SAT) involves determining the maximum number of clauses within a given Boolean formula in conjunctive normal form that can be satisfied simultaneously by assigning truth values to the variables. This problem extends beyond the scope of the Boolean satisfiability problem, which solely focuses on verifying the existence of a truth assignment satisfying all the clauses.

Let:
$n$: the number of variables
$m$: the number of clauses that have to be satisfied
$C$: the set of clauses
$S_i^p$: the set of variables that have to be true in the clause $i$
$S_i^n$: the set of variables that have to be false in the clause $i$
$w_i$: the weight of clause $i$

Variables
$y_i$: a binary variables that values 1 if the variables number $j$ value true
$z_i$: a binary variables that values 1 if the clause number $i$ value true

The objective consists in maximizing the weight of the satisfied clauses
$$\max \sum_{i=1}^{n} w_i . z_i$$
Subject to:
$$\forall i = 1, n \qquad z_i \leq \sum_{j \in S_i^p} y_j + \sum_{j \in S_i^n} (1 - y_j) \qquad (1)$$

A $k$-SAT problem can solve as a special case of the Max-k-SAT considering $w_i = 1$.

Quantum computing generalizes classical computing from binary bits to quantum bits, which may represent both 0's and 1's simultaneously in a superposition. SAT received attention of the Quantum community during the last years taking advantages of Grover Algorithm in (Paulet et al., 2023), (Lin et al., 2023) and (Varmantchaonala et al., 2023). The resolution is limited to small scale instances with, for example, 4 variables and 4 clauses in (Lin et al., 2023). (Yu et al., 2023) introduces an adaptive-bias QAOA (ab-QAOA) that solved the hard-region 3-SAT and Max-3-SAT problems effectively considering instances with 10 variables.

## 2. QAOA approach for SAT

The exploration of quantum approaches to optimization problems involves both modeling and implementation aspects in research. Among various domains, scheduling problems are particularly attractive for the application of quantum technology, given their inherently combinatorial nature. In a recent work by (Bourreau et al., 2023), a novel perspective on resolution was introduced. Instead of treating the problem as a Hamiltonian for function minimization, the focus shifted to the rank of solutions. The aim was to develop concise circuits with minimal gates suitable for current classical machines when addressing instances from literature. To achieve this, we adopt the concept of rank as defined by Laisant (1888), resulting in an approach termed IQAOA (Indirect QAOA) that extends the Quantum Approximate Optimization Algorithm (QAOA).

The indirect depiction of a solution leverages the inherent one-to-one correlation between permutations and a concept termed subexceedant functions. This approach significantly streamlines the representation of permutation ranks by employing a single integer. Numerous methods are available to establish this one-to-one correspondence, with the Lehmer code, commonly known as the inversion table, standing out as the most notable. The algorithmic description of this approach was initially introduced by Knuth in 1981 (Knuth, 1981).

In contrast to QAOA (Hadfield, 2018), which necessitates a Hamiltonian to define the function for minimization and may require a substantial number of gates, IQAOA simplifies the process by requiring only a Hamiltonian that models the ranks. This results in a highly compact circuit with a minimal number of gates and qubits. The IQAOA algorithm utilizes parametrized $\vec{\beta}$ and $\vec{\gamma}$ weights to define a quantum state $|\varphi(\vec{\beta},\vec{\gamma})\rangle$ representing a solution rank $x$. The probability of obtaining rank $x$, is given by $|\langle x|\varphi(\vec{\beta},\vec{\gamma})\rangle|^2$ and the expectation value is evaluated by sampling $\langle \varphi(\vec{\beta},\vec{\gamma})|C^p|\varphi(\vec{\beta},\vec{\gamma})\rangle$. Each generation provides a ranking within the solution list, and this rank can be translated into the corresponding solution. With fixed values for $\vec{\beta},\vec{\gamma}$, the quantum computer defines the state $|\varphi(\vec{\beta},\vec{\gamma})\rangle$, and a measurement in the computational basis produces a string $x$ for evaluating $\langle \varphi(\vec{\beta},\vec{\gamma})|C^p|\varphi(\vec{\beta},\vec{\gamma})\rangle$. The distinct advantage lies in the simultaneous manipulation of all ranks, effectively covering all potential solutions through the associated probability distribution. In contrast to conventional methods for solving problems, which often involve exploring a subset of the solution space, quantum approaches overcome the challenge by addressing the entire solution space. The complexity shifts towards modeling it as a Hamiltonian and determining the angles associated with the probability distribution (Bourreau et al., 2023).

The binary representation of rank is

$$\text{rank} = \sum_{j=0}^{n} x_j . 2^j \text{ with } x_j \in \{0; 1\} \tag{5}$$

with

$$H_P = \frac{1}{2}\sum_{j=0}^{n}(Id - Z_j).2^j \tag{6}$$

The algorithm principle is illustrated on Fig. 1. IQAOA efficiency relies on the following key-points:

- The capability to attain a favorable ratio in the estimation of $C^p\left(\vec{\beta},\vec{\gamma}\right)$ achieved through a meticulously calibrated number of necessary shots..

- The optimal distribution $|\psi(\vec{\beta^*},\vec{\gamma^*})\rangle$ once discovered, should be estimated from a restricted subset of solutions, taking into account the overall number of solutions. This approach helps circumvent the need for costly enumerations. Essentially, the algorithm needs to converge toward optimal and near-optimal solutions, reflecting that a substantial portion of probabilities is associated with these optimal or high-quality solutions.
- The presence of a specialized method for computing $(\vec{\beta^*},\vec{\gamma^*})$.
- The value of qubits, denoted as $p$, should satisfy $\geq n!$ and while keeping the quantum circuit's gate count low for encoding a permutation. This consideration is in relation to the classical QAOA approach and the corresponding circuit length

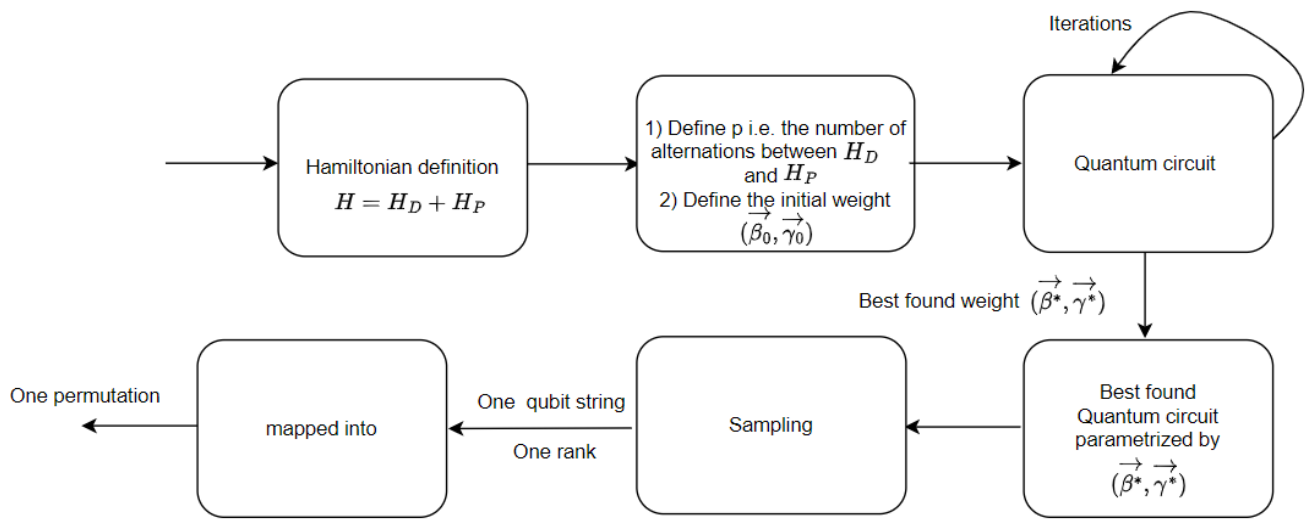

**Figure 1.** IQAOA principles (Bourreau et al., 2023)

The best parameters $(\vec{\beta^*},\vec{\gamma^*})$ can be computed any iterative based method including but limited to genetic algorithm, GRASP, simulated annealing.

In the very specific situation of the SAT, the binary string $(x_1, \ldots x_j, \ldots x_n)$ is the direct representation of one solution SAT solution and the binary string is fully defined by the measurement. By consequence a distribution $|\psi(\vec{\beta},\vec{\gamma})\rangle$ is a set of binary strings and each string is related to the number of shots experienced during the sampling.

The objective function, i.e. the cost associated to each binary string $(x_1, \ldots x_j, \ldots x_n)$ must be carefully defined to

- modelize the number of unsatisfied clauses;
- favor a probability distribution that converge during the generations on high-quality solutions, specifically solutions with no clauses violation.

Attaining the desired probability distribution involves choosing suitable minimization criteria. To construct a distribution that highlights probabilities associated with low-cost solutions, while avoiding an large number of values with residual probabilities, a possible approach is to incorporate a combination of both the mean and a criterion linked to the trend distribution, such as deciles or quartiles. This approach helps concentrate probabilities on cost-effective solutions while reducing residual probabilities across different values.

To favor the metaheuristic convergence, the appropriate criteria is defined as a weighted function:
$$g: [x] \to \mathbb{R}$$
$$g(x) = \varsigma . h(x) + \vartheta . d(x)$$
Where
$\varsigma$ and $\vartheta$: are two real number modeling the relative performance of $h(x)$ and $d(x)$ such that $\min \varsigma . h(x) > \max \vartheta . d(x)$ defining a strict hierarchy between the criteria
$h(x)$: the number of unsatisfied clauses in $x$
$d(x)$: the measure of divergence in the position of potentially non-satisfied clauses in $x$
$$d(x) = \sum_{i=1}^{n}(1 - z_i) . i^2$$
Let us note $cf_i$ the cumulated frequencies and $e_i$ the element related to $cf_i$ and let us define $e_p$ for $p \in ]0; 1[$:
$$e_p = j \text{ where } cf_j \geq p \text{ and } cf_{j-1} < p$$
For a $p \in ]0; 1[$, $e_p$ is the value that divide a population in such a way that $p \times 100$ percent of the population is above $e_p$. Let us define $E = \{e_{p_1}; e_{p_2}; e_{p_3}; \ldots ; e_{p_k}\}$ a set of $k$ distinct probabilities, for example, $E = \{e_{0.01}; e_{0.06}; e_{0.11}; e_{0.17}; e_{0.22}; e_{0.27}; e_{0.32}\}$ defines 7 values dividing the population with 1 percent above for $E_1 = e_{0.01}$, with 6 percents above for $E_2 = e_{0.06}$ and with 32 percents above for $E_7 = e_{0.32}$

Let us note $\bar{g}_s$ the estimator of the average $\langle \varphi(\vec{\beta}, \vec{\gamma}) | g(x) | \varphi(\vec{\beta}, \vec{\gamma}) \rangle$ using $s$ shots and $\bar{d}_s$ the estimator of the sum of $g(x)$ for $x \in E$:

$$C^p(\vec{\beta}, \vec{\gamma}) = \langle \varphi(\vec{\beta}, \vec{\gamma}) | g(x) | \varphi(\vec{\beta}, \vec{\gamma}) \rangle + \sum_{x \epsilon E} \langle \varphi(\vec{\beta}, \vec{\gamma}) | g(x) | \varphi(\vec{\beta}, \vec{\gamma}) \rangle$$

This function derives a probability distribution that prioritizes high-quality solutions, specifically targeting solutions with minimal unsatisfied clauses.

### 3. Numerical experiments

The experiments were carried out using the 91-clauses-20-variables instances in the DIMACS cnf format which supported by a large number of solvers available in the SATLIB solvers collection.
https://www.cs.ubc.ca/~hoos/SATLIB/benchm.html
The 1000 instances of the dataset are all satisfiable. Note that numerical experiments focus on this set of instances because the quantum simulator we use is limited to 32 qubits i.e. only 3-SAT with less than 32 variables can be solved using the simulator.

All the experiments have been carried out with:

- $p = 2$ i.e. a QAOA depth limited to 2
- A genetic algorithm implemented using the **pygad** python library.
- Each quantum state is evaluated using 250 shots

The genetic algorithm has been tuned considering:

- 150 generations;
- 30 solutions per population;
- A probability of 25% for mutation;
- A tournament parent selection
- 4 solutions are kept for elitism.

Be aware that the parameters were established following a brief numerical study and warrant special attention. All experiments were conducted utilizing Qiskit (IBM) and the simulator.

The preliminary experiments were carry out on a small-scale instance with 10 clauses and 5 variables.

$C_1 = x_1 \vee x_2 \vee x_3$
$C_2 = x_1 \vee x_2 \vee \bar{x}_3$
$C_3 = x_1 \vee \bar{x}_2 \vee x_3$
$C_4 = x_1 \vee \bar{x}_2 \vee \bar{x}_3$
$C_5 = \bar{x}_1 \vee x_2 \vee x_3$
$C_6 = \bar{x}_1 \vee x_2 \vee \bar{x}_3$
$C_7 = \bar{x}_1 \vee \bar{x}_2 \vee x_3$
$C_8 = x_1 \vee x_5 \vee x_4$
$C_8 = x_1 \vee x_5 \vee \bar{x}_4$
$C_{10} = x_1 \vee \bar{x}_5 \vee x_4$

This instance has 4 solutions only fully described below:
```
Solution 1 = [1, 1, 1, 0, 0]
Solution 2 = [1, 1, 1, 0, 1]
Solution 3 = [1, 1, 1, 1, 0]
Solution 4 = [1, 1, 1, 1, 1]
```
*i.e.*

solution 1. $x_1 = True, x_2 = True, x_3 = True$
solution 1. $x_1 = True, x_2 = True, x_3 = True, x_5 = True$
solution 1. $x_1 = True, x_2 = True, x_3 = True, x_4 = True$
solution 1. $x_1 = True, x_2 = True, x_3 = True, x_4 = True, x_5 = True$

The total number of permutations is 32 and there is only 3 different values of $h(x)$: there is only 4 permutation leading to a feasible solution, 16 permutations where one constraint is not satisfied and the larger number of unsatisfied constrained in 2 (table 1).

**Table 1.** Initial distribution of solution considering $h(x)$ v.s. final distribution

| $h(x)$ | Initial distribution 32 solutions | | Final distribution 100 000 samplings | |
|---|---|---|---|---|
| | Number of solutions | Probability | Number of solutions | Probability |
| 0 | 4 | 12.500 | 89173 | 89.173 |
| 1 | 16 | 50.000 | 10725 | 10.725 |
| 2 | 12 | 37.500 | 102 | 0.102 |
| 3 | 0 | 0.000 | 0 | 0.000 |

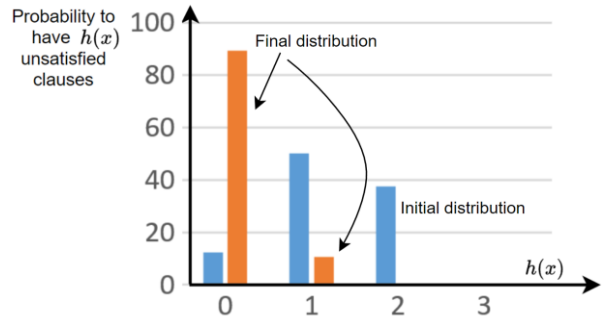

**Figure 2.** Visualization of the two distribution for the 10-clause and 5-variable instance

After optimization, the probability to obtain a solution satisfying all the clauses is about 89% proving the efficiency of the approach in this widget example (Fig. 2).

The first instance referred to as **uf20-01.cnf** is composed of 20 variables and 91 clauses and 8 solutions satisfy the 91 clauses:
```
Solution 1 = [0, 1, 1, 1, 0, 0, 0, 1, 1, 1, 1, 0, 0, 1, 1, 0, 1, 1, 1, 1]
Solution 2 = [1, 0, 0, 0, 0, 1, 0, 0, 0, 0, 0, 0, 1, 1, 1, 0, 1, 0, 0, 1]
Solution 3 = [1, 0, 0, 0, 0, 1, 0, 0, 1, 0, 0, 0, 0, 1, 1, 0, 1, 0, 0, 1]
Solution 4 = [1, 0, 0, 0, 0, 1, 0, 0, 1, 0, 0, 0, 1, 1, 1, 0, 1, 0, 0, 1]
Solution 5 = [1, 0, 0, 1, 0, 0, 0, 0, 0, 1, 0, 0, 1, 1, 1, 0, 1, 0, 0, 1]
Solution 6 = [1, 0, 0, 1, 0, 0, 0, 1, 0, 1, 0, 0, 1, 1, 1, 0, 1, 0, 0, 1]
Solution 7 = [1, 0, 0, 1, 0, 1, 0, 0, 0, 0, 0, 0, 1, 1, 1, 0, 1, 0, 0, 1]
Solution 8 = [1, 0, 0, 1, 0, 1, 0, 0, 0, 1, 0, 0, 1, 1, 1, 0, 1, 0, 0, 1]
```
The total number of permutations is 1 048 576 but there is only 29 different value of $h(x)$: there is only 8 permutation leading to a feasible solution, 82 permutations where one constraint is not satisfied and the larger number of unsatisfied constrained is 29.

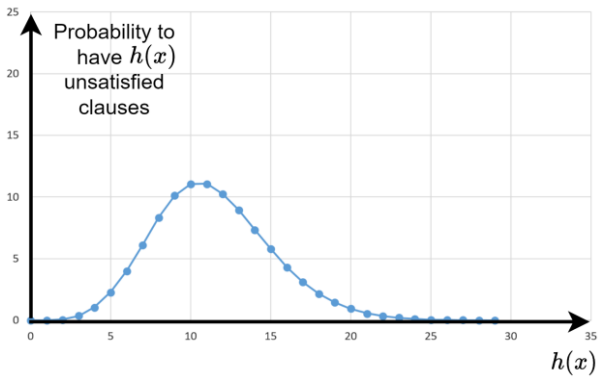

**Figure 3.** Initial distribution of solutions considering $h(x)$ - uf20-01.cnf

A enumeration of permutations permits to prove that the high quality solutions have a very low probability (Fig. 3): the probability to obtain one solution (the 91 constraints are satisfied) is about 0.0007% and the higher probabilities are related to assignment with about 11 unsatisfied constraints.

**Table 2.** Initial distribution of solution considering $h(x)$

| $h(x)$ | Number of solutions | Probability |
|---|---|---|
| 0 | 8 | 0.000763 |
| 1 | 82 | 0.007820 |
| 2 | 648 | 0.061798 |
| 3 | 3745 | 0.357151 |
| 4 | 11047 | 1.053524 |
| 5 | 23921 | 2.281284 |
| 6 | 42032 | 4.008484 |
| 7 | 64051 | 6.108379 |
| 8 | 87545 | 8.348942 |
| 9 | 105946 | 10.103798 |
| 10 | 115744 | 11.038208 |
| 11 | 116202 | 11.081886 |
| 12 | 107416 | 10.243988 |
| 13 | 93822 | 8.947563 |
| 14 | 77199 | 7.362270 |
| 15 | 60595 | 5.778790 |
| 16 | 44949 | 4.286671 |
| 17 | 32450 | 3.094673 |
| 18 | 22658 | 2.160835 |
| 19 | 15100 | 1.440048 |
| 20 | 9811 | 0.935650 |
| 21 | 6022 | 0.574303 |
| 22 | 3491 | 0.332928 |
| 23 | 2110 | 0.201225 |
| 24 | 1056 | 0.100708 |
| 25 | 528 | 0.050354 |
| 26 | 260 | 0.024796 |
| 27 | 93 | 0.008869 |
| 28 | 40 | 0.003815 |
| 29 | 5 | 0.000477 |

The decile is 7 (with about 13% of the cumulated frequencies) i.e. we have a probability about 13% to have an assignment with less than 7 unsatisfied clauses. The cumulative frequency proves that we have a probability of 54% to have less than 11 clauses as stressed on table 2.

The genetic algorithm is used for optimization with $E = \{e_{0.01}; e_{0.05}; e_{0.1}\}$ considering the set of parameters introduced at the beginning of section 3.

**Table 3.** Final distribution

| $h(x)$ | Number of solutions | Probability |
|---|---|---|
| 0 | 243 | 0.243 |
| 1 | 4015 | 4.015 |
| 2 | 19266 | 19.266 |
| 3 | 12770 | 12.770 |
| 4 | 11299 | 11.299 |
| 5 | 11965 | 11.965 |
| 6 | 9041 | 9.041 |
| 7 | 7580 | 7.580 |
| 8 | 6863 | 6.863 |
| 9 | 6269 | 6.269 |
| 10 | 3660 | 3.660 |
| 11 | 2902 | 2.902 |
| 12 | 1866 | 1.866 |
| 13 | 1072 | 1.072 |
| 14 | 646 | 0.646 |
| 15 | 310 | 0.310 |
| 16 | 129 | 0.129 |
| 17 | 60 | 0.060 |
| 18 | 33 | 0.033 |
| 19 | 7 | 0.007 |
| 20 | 3 | 0.003 |
| 21 | 1 | 0.001 |
| 30 | 0 | 0.000 |

The best found distribution $|\psi(\vec{\beta^*},\vec{\gamma^*})\rangle$ is estimated by sampling with 100 000 shots giving the final distribution of Fig. 4.

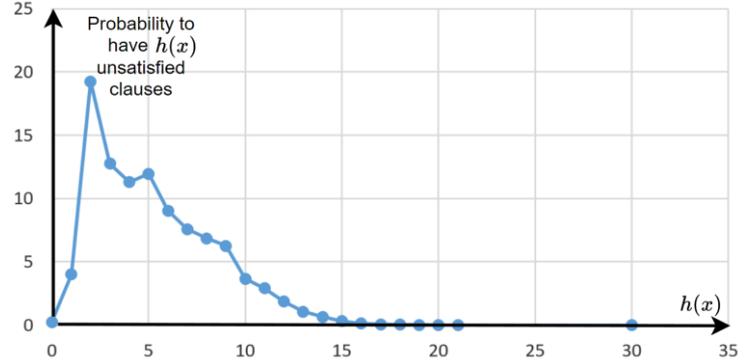

**Figure 4.** Final distribution of solutions considering $h(x)$ – instance uf20-01.cnf

The information presented in Table 3 corroborates the observations from the visual representation, indicating notably high probabilities focused on low-cost solutions. This serves to underscore the effectiveness of the IQAOA method in resolving instances.

The probability to obtain a solution satisfying all the clauses is about 0.243% that is 318 times better than the 0.0007% of the initial distribution. Let us note that the decile is now 2 and we have a probability about 59% to have a solution with less than 5 unsatisfied clauses.

The second instance, identified as **uf20-02.cnf**, consists of 20 variables and 91 clauses, with 29 solutions satisfying all 91 clauses. The depictions of the distributions in Fig. 5 and Table 4 reveal a noteworthy shift in the probability distribution toward the left, now concentrating on high-quality solutions.

It should be noted in table 4 that, in the initial distribution, the median is 11 in the initial distribution, which means that 50% of the data corresponds to solutions with less than 11 unsatisfied clauses. The final sampling achieved at the end of the optimization gives:

- a median that values 6;
- A probability of 1.698% to satisfied all the clauses that is 613 times better than the 0.003% of the initial distribution.

**Table 4.** Initial and final distributions for instance uf20-02.cnf

| | Probabilities | |
|---|---|---|
| $h(x)$ | Initial distribution | Final distribution |
| 0 | 0.003 | 1.698 |
| 1 | 0.021 | 2.757 |
| 2 | 0.085 | 7.315 |
| 3 | 0.291 | 8.797 |
| 4 | 0.769 | 11.727 |
| 5 | 1.679 | 11.528 |
| 6 | 3.246 | 12.155 |
| 7 | 5.394 | 11.538 |
| 8 | 7.797 | 9.275 |
| 9 | 10.055 | 7.771 |
| 10 | 11.653 | 5.671 |
| 11 | 12.228 | 3.953 |
| 12 | 11.578 | 2.578 |
| 13 | 10.104 | 1.522 |
| 14 | 8.203 | 0.822 |
| 15 | 6.095 | 0.489 |
| 16 | 4.242 | 0.194 |
| 17 | 2.786 | 0.109 |
| 18 | 1.717 | 0.047 |
| 19 | 0.979 | 0.036 |
| 20 | 0.532 | 0.009 |
| 21 | 0.279 | 0.002 |
| 22 | 0.146 | 0.004 |
| 23 | 0.069 | 0.003 |
| 24 | 0.031 | |
| 25 | 0.012 | |
| 26 | 0.004 | |
| 27 | 0.002 | |
| 28 | 0.001 | |
| 29 | 0.000 | |

The fact that the probability distribution only marks certain solutions, and that this distribution is very different from a uniform distribution, is clearly visible in the diagram in figure 5.

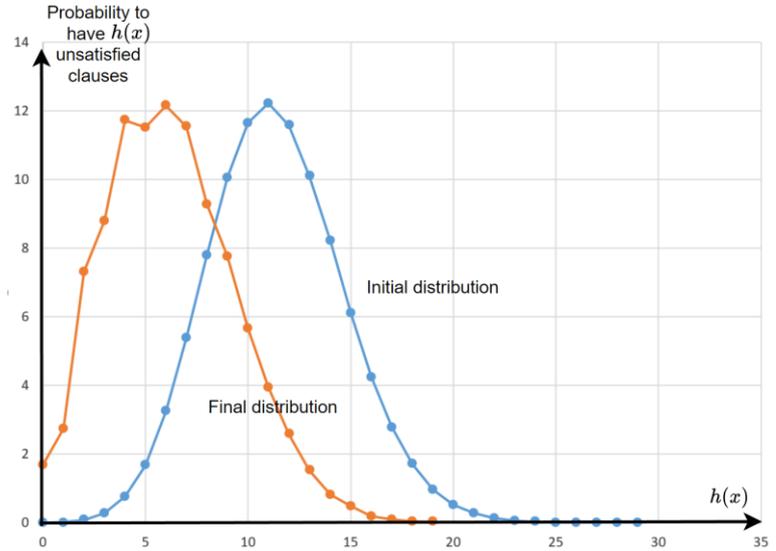

**Figure 5.** Comparison of initial and final distribution considering $h(x)$ – instance uf20-02.cnf

For the instance number 3 (uf20-03.cnf), there is only one assignment that permit to satisfy all the clauses (table 5) and the optimization permits to obtain a probability of 0.278% for $h(x) = 0$ and a decile that value 3 (table 5).

It's worth mentioning that $\sum_{x \in E} \langle \varphi(\vec{\beta}, \vec{\gamma}) | g(x) | \varphi(\vec{\beta}, \vec{\gamma}) \rangle$ s crucial for evaluating the distribution and variability of data, helping to identify potential outliers and extreme values.

**Table 5.** Initial and final distributions for instance uf20-03.cnf

| $h(x)$ | Probabilities in the final distribution |
|---|---|
| 0 | 0.278 |
| 1 | 1.556 |
| 2 | 4.923 |
| 3 | 6.881 |
| 4 | 14.119 |
| 5 | 14.965 |
| 6 | 16.155 |
| 7 | 14.583 |
| 8 | 10.234 |
| 9 | 7.173 |
| 10 | 4.215 |
| 11 | 2.137 |
| 12 | 1.535 |
| 13 | 0.721 |
| 14 | 0.365 |
| 15 | 0.105 |
| 16 | 0.04 |
| 17 | 0.008 |
| 18 | 0.004 |
| 19 | 0.003 |
| 29 | 0.000 |

To obtain a probability distribution that concentrates probabilities on low-$h(x)$ solutions and avoids having a large number of values with residual probabilities we have to consider combinations of both the mean and a criterion related to the mean trend.

Definitions of $E = \{e_{p_1}; e_{p_2}; e_{p_3}; ...; e_{p_k}\}$ with a large-enough $k$ distinct probabilities, used to model the left-part of the distribution, favor concentration of probabilities on high-quality assignment.

A numerical test conducted and presented hereafter push us into considering that $E = \{e_{p_1}; e_{p_2}; e_{p_3}; ...; e_{p_k}\}$ must be tuned depending on both instances and on the probabilities that is expected on high-quality solutions.

The experiments to highlight the importance of $E$ in the final probabilities concentration on high quality solutions can be evaluated considering:

- the instance number 4 (uf20-04.cnf);
- a number of generations fixed to 500 for the Genetic Algorithm;
- the consecutive resolution with 4 set:
    1. $E_1 = \{e_{0.01}; e_{0.06}\}$
    2. $E_2 = \{e_{0.01}; e_{0.06}; e_{0.11}\}$
    3. $E_3 = \{e_{0.01}; e_{0.06}; e_{0.11}; e_{0.16}; e_{0.21}\}$
    4. $E_4 = \{e_{0.01}; e_{0.06}; e_{0.11}; e_{0.16}; e_{0.21}, e_{0.26}, e_{0.32}\}$

For the instance number 4 there is only 3 assignments satisfying the clauses:
```
Solution 1 = [1, 0, 1, 1, 0, 0, 0, 0, 0, 1, 0, 0, 1, 0, 0, 1, 1, 0, 0, 0]
Solution 2 = [1, 0, 1, 1, 0, 0, 1, 0, 0, 1, 0, 0, 1, 0, 0, 1, 1, 0, 0, 0]
Solution 3 = [1, 0, 1, 1, 0, 0, 1, 0, 0, 1, 1, 0, 1, 0, 0, 1, 1, 0, 0, 0]
```

The probability assigned to $h(x) = 0$ is about 0.0003%, the decile is 7 meaning that we have a probability of no more that 10% to have an assignment leading to less than 7 unsatisfied clauses and the median is 11. Much like the earlier instance, a sampling of permutations yields a comparable conclusion (refer to Table 6).

**Table 6.** Initial and final distributions for instance uf20-04.cnf depending on $E$

| | Probabilities in | | | | |
|---|---|---|---|---|---|
| | **Initial distribution** | **final distribution** | | | |
| $h(x)$ | | $E_1$ | $E_2$ | $E_3$ | $E_4$ |
| 0 | 0.0003 | 0.012 | 0.045 | 0.983 | 1.226 |
| 1 | 0.0022 | 0.074 | 0.345 | 1.744 | 2.169 |
| 2 | 0.0278 | 1.819 | 1.792 | 12.604 | 5.24 |
| 3 | 0.1759 | 7.184 | 4.088 | 16.125 | 6.89 |
| 4 | 0.5805 | 16.99 | 7.352 | 19.043 | 8.987 |
| 5 | 1.4780 | 15.737 | 12.618 | 15.127 | 12.864 |
| 6 | 3.0790 | 15.623 | 17.594 | 13.053 | 13.8 |
| 7 | 5.4389 | 13.188 | 19.492 | 9.315 | 15.417 |
| 8 | 8.1277 | 11.535 | 16.848 | 4.908 | 13.087 |
| 9 | 10.4388 | 7.553 | 9.649 | 3.411 | 9.554 |
| 10 | 11.7940 | 4.769 | 5.445 | 2.075 | 5.683 |
| 11 | 12.1699 | 2.606 | 2.742 | 0.939 | 2.906 |
| 12 | 11.5840 | 1.437 | 1.228 | 0.434 | 1.224 |
| 13 | 10.1158 | 0.835 | 0.465 | 0.169 | 0.577 |
| 14 | 8.2635 | 0.387 | 0.19 | 0.043 | 0.236 |
| 15 | 6.1821 | 0.144 | 0.074 | 0.014 | 0.095 |
| 16 | 4.3645 | 0.064 | 0.022 | 0.005 | 0.032 |
| 17 | 2.8523 | 0.031 | 0.011 | 0.006 | 0.007 |
| 18 | 1.6858 | 0.009 | | 0.002 | 0.005 |
| 19 | 0.9212 | 0.003 | | 0.983 | |
| 20 | 0.4327 | | | | |
| 21 | 0.1842 | | | | |
| 22 | 0.0738 | | | | |
| 23 | 0.0216 | | | | |
| 24 | 0.0046 | | | | |
| 25 | 0.0008 | | | | |
| 26 | 0.0001 | | | | |
| 27 | 0.0000 | | | | |

The numerical experiments proposed on Table 6 demonstrate the significance of parameter $E$ in the objective function, which allows for "controlling" the shape of the probability distribution.

A larger large-enough number of distinct probabilities in $E$, that is composed of low-level value (strictly lower that 0.5), increase the relative weigh of $\sum_{x \in E}\langle\varphi(\vec{\beta},\vec{\gamma})|g(x)|\varphi(\vec{\beta},\vec{\gamma})\rangle$ to $\langle\varphi(\vec{\beta},\vec{\gamma})|g(x)|\varphi(\vec{\beta},\vec{\gamma})\rangle$ and favor concentration on high-quality assignment i.e. concentration on assignment with very few unsatisfied clauses. The left-shift of distributions and the concentration on quality solutions is emphasized, for instance, in Fig. 6.

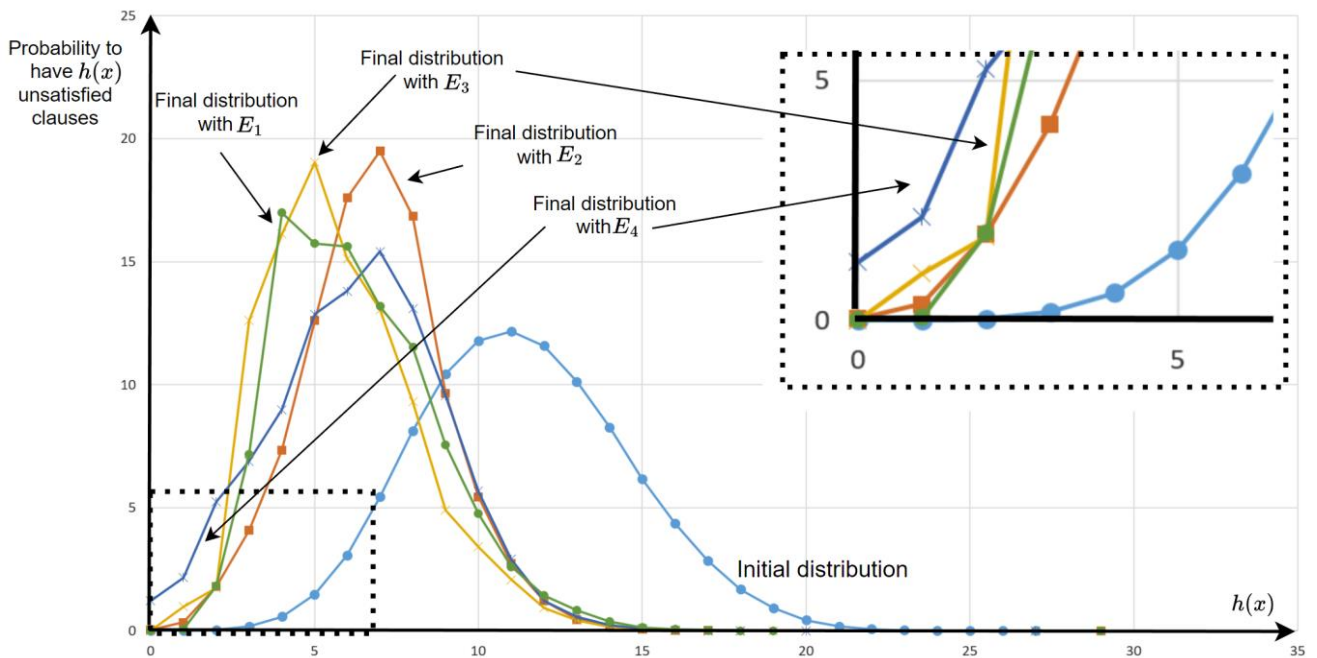

**Figure 6.** Comparison of final distribution according to $E$

**Table 7.** Final distributions for instance uf20-04.cnf with $E_4$ depending on the number of generations

| $h(x)$ | 500 generations | 1000 generations |
|---|---|---|
| 0 | 1.226 | 3.243 |
| 1 | 2.169 | 9.422 |
| 2 | 5.240 | 19.996 |
| 3 | 6.890 | 17.314 |
| 4 | 8.987 | 11.344 |
| 5 | 12.864 | 8.875 |
| 6 | 13.800 | 10.543 |
| 7 | 15.417 | 7.788 |
| 8 | 13.087 | 4.554 |
| 9 | 9.554 | 2.640 |
| 10 | 5.683 | 1.942 |
| 11 | 2.906 | 1.095 |
| 12 | 1.224 | 0.719 |
| 13 | 0.577 | 0.236 |
| 14 | 0.236 | 0.138 |
| 15 | 0.095 | 0.095 |
| 16 | 0.032 | 0.032 |
| 17 | 0.007 | 0.014 |
| 18 | 0.005 | 0.008 |
| 19 | 0.001 | 0.002 |
| 20 | 0.000 | 0.000 |

These results prove that the IQAOA method succeeds into transforming the amplitude of an initial distribution into that of a target state. However, very large $E$ should require more iterations of the method, since we can obtain objective function where $\sum_{x \in E} \langle \varphi(\vec{\beta}, \vec{\gamma}) | g(x) | \varphi(\vec{\beta}, \vec{\gamma}) \rangle$ that model the trend distribution on too large number of values, lies to objective function more complex to minimize.

This remark can be illustrated on table 7 considering $E_4 = \{e_{0.01}; e_{0.06}; e_{0.11}; e_{0.16}; e_{0.21}, e_{0.26}, e_{0.32}\}$ that is composed of 7 probabilities and Fig. 7 gives a graphical representation to highlight the impact of the number of generation.

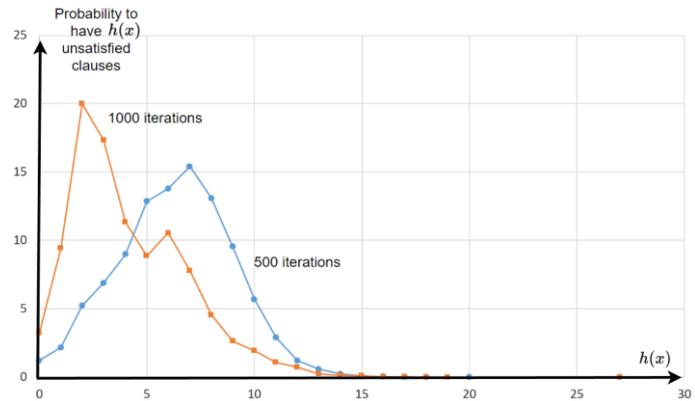

**Figure 7.** Importance of $E$ related to the number of generations

The results proves that after 500 generations only the distribution is not yet concentrated on assignments with $h(x) \in \{0; 1\}$ and that 1000 generations are required to obtain a probability distribution concentrating on high-quality solutions (Fig. 7.).

## 4. Concluding remarks

This paper explores the application of the IQAOA approach to tackle the 3-SAT problem. The IQAOA method relies on an indirect representation of solutions through highly compact circuits with a minimal number of gates.

To achieve a probability distribution emphasizing high-quality solutions, particularly those with minimal unsatisfied clauses, the approach employs: 1) a dedicated objective function that combines both the mean and a criterion associated with the trend distribution; and 2) a genetic algorithm for determining the optimal parameters $(\overrightarrow{\beta^*}, \overrightarrow{\gamma^*})$.

Evaluating the effectiveness of this method on 91-clauses and 12-variables instances showcases a probability distribution that concentrate on the solution that minimize the unsatisfied number of clauses. To the best of our knowledge, this represents the first quantum resolution to such large 3-SAT instances.